\documentclass[12pt]{article}
\usepackage[a4paper, top=0.8in, bottom=1in, left=1in, right=1in]{geometry}
\usepackage{amsmath, amssymb}
\usepackage{graphicx}
\usepackage{titlesec}
\usepackage{setspace}
\usepackage{url}
\usepackage{hyperref}
\usepackage{cite}
\usepackage{enumitem}
\usepackage{titling}
\usepackage{float}
\usepackage{caption}
\usepackage{xcolor}
\usepackage{indentfirst}
\setlist{noitemsep} 
\DeclareCaptionLabelSeparator{bar}{ \textbar\ }

\title{\textbf{Knowledge Synthesis of Photosynthesis Research Using a Large Language Model}\vspace{-0.5ex}}
\author{
\begin{minipage}{\textwidth}
\hspace*{0.5cm}  
\centering
\setstretch{1.2}
Seungri Yoon$^{1}$, Woosang Jeon$^{2}$, Sanghyeok Choi$^{2}$, Taehyeong Kim$^{2}$\footnotemark[1], Tae In Ahn$^{1,3}$\footnotemark[1]\\[0.5ex]
{\small
$^{1}$Department of Agriculture, Forestry and Bioresources, Seoul National University, Seoul 08826, Republic of Korea\\[0.2ex]
$^{2}$Department of Biosystems Engineering, Seoul National University, Seoul 08826, Republic of Korea\\[0.2ex]
$^{3}$Research Institute of Agriculture and Life Sciences, Seoul National University, Seoul 08826, Republic of Korea\\[0.2ex]
}
\end{minipage}
}

\date{}

\begin{document}
\raggedbottom

\maketitle

\renewcommand{\thefootnote}{\fnsymbol{footnote}}
\footnotetext[1]{Corresponding authors: \{taehyeong.kim, tiahn\}@snu.ac.kr}

\begin{abstract}
The development of biological data analysis tools and large language models (LLMs) has opened up new possibilities for utilizing AI in plant science research, with the potential to contribute significantly to knowledge integration and research gap identification. Nonetheless, current LLMs struggle to handle complex biological data and theoretical models in photosynthesis research and often fail to provide accurate scientific contexts. Therefore, this study proposed a photosynthesis research assistant (PRAG) based on OpenAI's GPT-4o with retrieval-augmented generation (RAG) techniques and prompt optimization. Vector databases and an automated feedback loop were used in the prompt optimization process to enhance the accuracy and relevance of the responses to photosynthesis-related queries. PRAG showed an average improvement of 8.7\% across five metrics related to scientific writing, with a 25.4\% increase in source transparency. Additionally, its scientific depth and domain coverage were comparable to those of photosynthesis research papers. A knowledge graph was used to structure PRAG's responses with papers within and outside the database, which allowed PRAG to match key entities with 63\% and 39.5\% of the database and test papers, respectively. PRAG can be applied for photosynthesis research and broader plant science domains, paving the way for more in-depth data analysis and predictive capabilities.
\end{abstract}

\section{Introduction}
Scientific research is a sophisticated process in which facts, concepts, and hypotheses interact intricately to create knowledge. Researchers have reviewed extensive academic literature and synthesized existing phenomena and theories using mathematical and logical concepts \cite{ref1}. Although these tasks are essential for deriving insights and hypotheses, the cognitive burden of researchers has increased, particularly as interdisciplinary research expands \cite{ref1, ref2, ref3, ref4}. These limitations are particularly evident in modern scientific research, which involves complex data processing and information overload.

In this context, large language models (LLMs) have shown the potential to support researchers' cognitive processes by partially simulating human language understanding and reasoning \cite{ref5, ref6, ref7}. However, academic reviews on the extent to which LLMs can replace or support scientific thinking remain insufficient, with notable limitations in information accuracy and source transparency \cite{ref8, ref9, ref10}.

Photosynthesis research is a representative example of the complexity and cognitive burden of modern scientific research. Photosynthesis is a key process that maintains ecosystem stability by converting solar energy into chemical energy, thereby directly impacting climate change and food security \cite{ref11, ref12, ref13}. Research in this field spans a wide range of disciplines, including biology, chemistry, physics, agricultural science, and environmental science, and the volume of available data has increased exponentially over recent decades \cite{ref14, ref15, ref16, ref17}. However, owing to information overload, traditional analytical methods face challenges in efficiently processing large volumes of data and deriving significant insights \cite{ref4, ref18, ref19}.

To address these challenges, we proposed a specialized photosynthesis research assistant (PRAG) by employing OpenAI's GPT-4o model with the retrieval-augmented generation (RAG) technique with prompt optimization to a database of photosynthesis research papers (Fig. 1). PRAG ensures transparency in citations based on actual research papers and is specialized for photosynthesis research. For instance, PRAG demonstrated an average improvement of 8.7\% over the baseline GPT-4o model across five metrics (scientific accuracy, research goal fit, source transparency, academic tone, and information reliability) required for scientific paper writing, with a notable 25.4\% improvement in source transparency. Although PRAG used only 5\% of the text from papers in the cited literature, its responses to photosynthesis research paper hypotheses reached or exceeded the levels of scientific depth (102.3\%) and domain coverage (99.3\%), thus reflecting the enrichment of knowledge and indicating the academic scope of the discussion. Even for papers not included in the database, PRAG matched approximately 39.5\% of the key entities of test papers, and an analysis of the spatiotemporal distribution of concepts showed a high correlation between the two datasets, with $R^2 = 0.65$ (test paper’s entities) and $0.61$ (PRAG’s entities).
\begin{figure}[h!]
    \centering
    \includegraphics[width=1\textwidth]{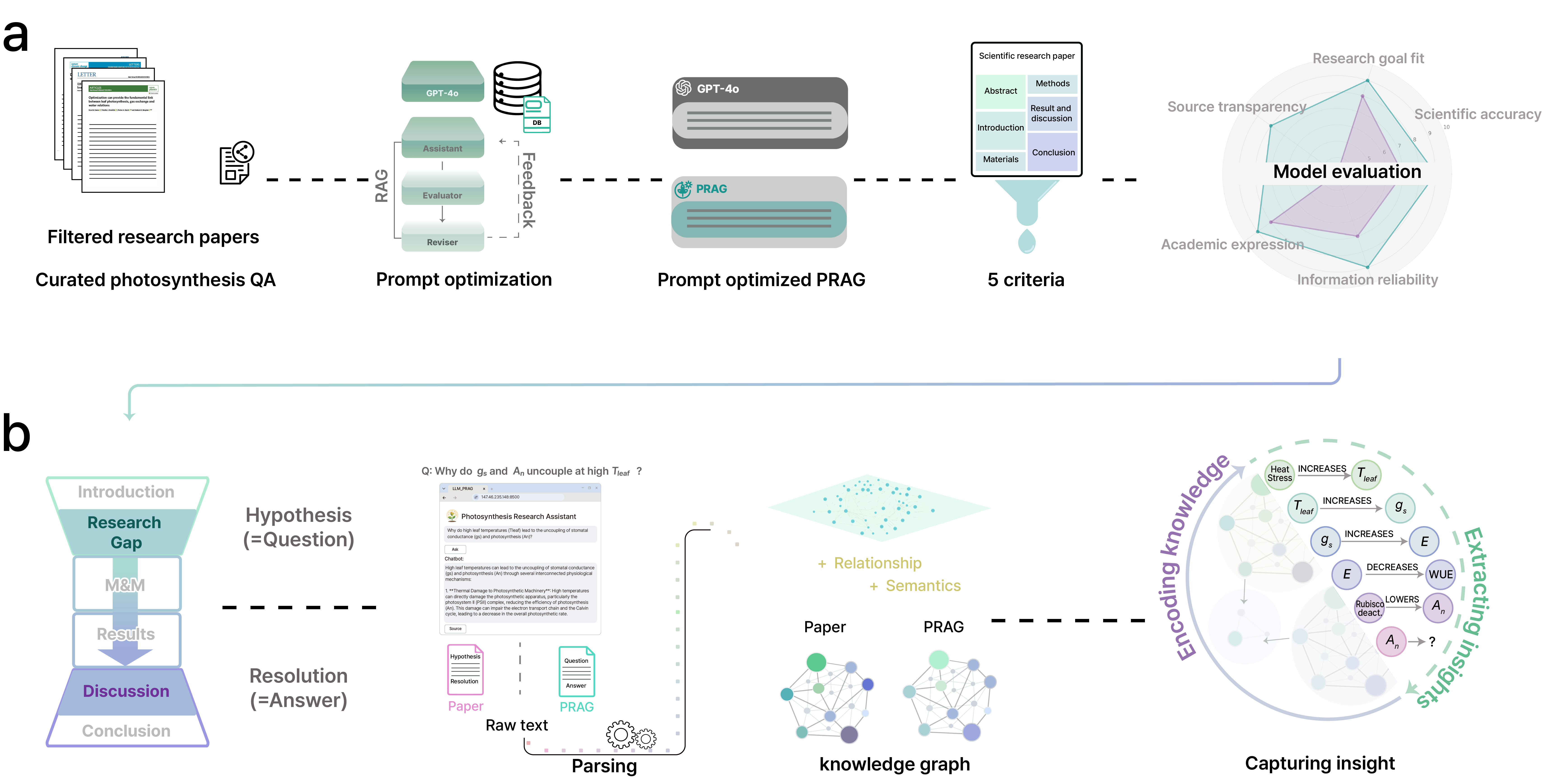}
    \captionsetup{
    labelfont=bf,
    labelsep=bar,
    font={small}
}
    \caption{Overview of our contributions. a, We selected 150 photosynthesis research papers with high citation indices to generate and curate research questions and answers (QAs). Then, the papers were incorporated into a vector database (DB), and the GPT-4o model was improved using RAG and prompt optimization processes. PRAG and the baseline GPT-4o model were comparatively evaluated based on five metrics required for scientific paper writing. b, Core components of scientific papers, such as hypotheses and discussions, were mapped to the LLM's question-and-answer format. We prepared database papers and test papers, generated PRAG responses to the hypotheses of each paper in raw text, and structured and parsed these responses for visualization in a knowledge graph. Finally, we evaluated PRAG's potential to provide scientific insights through structural and semantic comparative analyses.}
    \label{fig:prag}
\end{figure}

The photosynthesis-specialized PRAG can inspire researchers by proposing hypotheses, partially performing the scientific discovery cycle, and linking these hypotheses to predictions or discussions. To illustrate this function, we utilized a knowledge graph to organize and compare PRAG's responses to hypotheses from papers within and outside the database (Fig. 1b). For instance, PRAG elaborated on the impact of far-red photons on C3 and C4 photosynthesis at the molecular level through cyclic electron flow, expanding the discussion to areas such as agricultural and environmental dimensions, including crop productivity and ecosystem photosynthesis.

This study has demonstrated that enhancing existing LLMs with RAG techniques and prompt optimization can effectively integrate and expand the extensive academic literature in the field of photosynthesis, thereby alleviating researchers' cognitive burden and supporting the scientific reasoning process.

\section{Methods}
\subsection{Model Enhancement}
To improve GPT-4o to support photosynthesis research, we used a prompt optimization approach that combined RAG techniques with a self-feedback mechanism. The main goal was to address the limitations of existing LLMs in providing accurate and relevant responses to complex scientific questions related to photosynthesis.

\subsection{Pipeline Development}
This study was conducted using OpenAI's GPT-4o model as the backbone, which is characterized by strong inferential capabilities and appropriate citations based on web searches. We integrated RAG techniques to enhance the accessibility to external databases and provide more accurate and enriched information. Additionally, to develop a photosynthesis research assistant capable of providing scientifically accurate and reliable responses, we used key prompt strategies such as role assignment, elaboration, context provision, examples, and guideline definition.

\begin{itemize}
    \item \textbf{RAG Assistant}: RAG Assistant uses the RAG model to search for relevant documents or data from external databases and generates responses using that context. The model was trained to understand queries and integrate relevant information to generate consistent answers.
    \item \textbf{RAG Evaluator}: RAG Evaluator assesses the quality of the responses generated by RAG Assistant based on five criteria.
    \begin{enumerate}
        \item Scientific accuracy
        \item Research goal fit
        \item Source transparency
        \item Academic tone
        \item Information reliability
    \end{enumerate}
    \item \textbf{Prompt Reviser}: Prompt Reviser modifies prompts based on the feedback provided by RAG Evaluator. Specifically, it focuses on improving responses rated below Q1 by analyzing prompts that lead to low-quality responses and restructuring them more effectively to draw better information from the retrieval system or model.
\end{itemize}

\subsection{System Prompt}
Please refer to Appendix (SI. T1) for the system prompts and optimized prompts for each PRAG submodule.

\subsection{Research Text Visualization}
\textbf{Automated PDF Evaluation}: To map and visualize photosynthesis research texts across various scales, including academic domains as well as the spatiotemporal aspects of plant systems, we developed an automated PDF evaluation tool to process large numbers of PDF files using the OpenAI GPT-4o-mini model. This tool reads the PDF files, splits the text, and evaluates the text based on the scientific depth and domain coverage scores. Through system prompts (SI. T4), the evaluation model iterates over all PDF files in a folder, analyzing specific sections, such as the Discussion or Main text sections.

\subsection{Knowledge Graph Construction}
\textbf{PDF Parser}: Using the PyMuPDF library in Python, useful information was extracted from unstructured text data. The extracted text was chunked into specific formats for analysis, and concepts and entities were extracted using the OpenAI GPT-4o-mini model. This process is based on natural language processing (NLP) techniques, such as entity recognition and relationship extraction \cite{ref26}.

\textbf{NetworkX Module}: NetworkX is a library used in Python to create and analyze graph data and thus is useful for visualizing knowledge graphs. Using NetworkX, we visualized the knowledge graph by representing entities extracted during the parsing process as nodes and their relationships as edges. This allowed us to intuitively represent various concepts and their relationships extracted from photosynthesis research and PRAG discussions.

\subsection{Dataset}
To evaluate the potential of LLMs in photosynthetic research, we focused on answering questions related to photosynthesis. Answering these questions requires an understanding of plant physiological knowledge and plant-nature interactions across a wide range of fields.

The photosynthesis QA constructed in this study included both a dataset related to interdisciplinary photosynthesis research containing 2--3-sentence answers and a longer dataset that includes the logical development of scientific papers. An additional 2000 sets were selected for training and testing by filtering QAs that contained specific information available only through scientific papers. All datasets were provided in English.

\textbf{Trainset}: A total of 150 questions generated based on papers published within the past five years were used to optimize the prompts.

\textbf{Test set}: Final performance evaluation of 2000 questions published after 2021. The overall performance distribution was assessed using prompts.

\subsection{QA Examples by Research Domains from Photosynthesis Papers in the DB}
\textbf{Genetic Engineering} \cite{ref31}
\begin{itemize}
    \item Example question: What is the HYR gene?
    \item Answer: The Higher Yield Rice (HYR) gene refers to a transcription factor (TF) known to increase rice yield. It is part of the AP2/ERF (APETALA2/ethylene-responsive element binding factor) domain-containing TF family and enhances yield under drought, improves photosynthetic capacity, promotes strong root system development, and leads to higher water use efficiency among other positive physiological traits.
\end{itemize}

\textbf{Plant Physiology/Molecular Physiology} \cite{ref32}
\begin{itemize}
    \item Example question: Are the regulatory patterns of suberin biosynthesis conserved across monocot and dicot lineages?
    \item Answer: Yes, the regulatory patterns of suberin biosynthesis are conserved across monocot and dicot lineages. This is suggested by the co-expression of regulatory candidates with biosynthesis gene targets in both maize and Arabidopsis thaliana, indicating strong conservation in suberin developmental programs.
\end{itemize}

\textbf{Plant Physiology} \cite{ref33}
\begin{itemize}
    \item Example question: What limits photosynthetic assimilation in stressed leaves?
    \item Answer: Water stress limits photosynthetic assimilation primarily by inhibiting ATP synthesis due to decreased ATP synthase content, which in turn reduces the supply of ribulose bisphosphate (RuBP) necessary for the Calvin cycle. This inhibition is not due to impaired CO$_2$ diffusion or photoinhibition but is directly linked to metabolic constraints imposed by reduced ATP availability.
\end{itemize}

\textbf{Biochemistry/Molecular Biology} \cite{ref34}
\begin{itemize}
    \item Example question: What is the primary function of cytochrome bc complexes?
    \item Answer: The primary function of cytochrome bc complexes is to couple the redox energy of electron transfer reactions to proton translocation across the membranes of bacteria, mitochondria, and chloroplasts. This process conserves energy from the oxidation-reduction reactions in the form of an electrochemical proton gradient that can be used to synthesize ATP, transport solutes, or perform other work.
\end{itemize}

\textbf{Biochemistry/Plant Physiology} \cite{ref35}
\begin{itemize}
    \item Example question: What is the potential role of aquaporins in leaf hydraulic conductance?
    \item Answer: Aquaporins facilitate water transport across cell membranes and play a significant role in regulating leaf hydraulic conductance by enhancing the efficiency of water movement through leaf tissues. This regulation is crucial for maintaining adequate water supply to the mesophyll and optimizing stomatal aperture and gas exchange, particularly under varying environmental conditions, such as light and dehydration.
\end{itemize}

\textbf{Structural Biology} \cite{ref36}
\begin{itemize}
    \item Example question: How does leaf anatomy influence photosynthesis?
    \item Answer: Leaf anatomy significantly influences photosynthesis by determining the distribution and efficiency of light capture and CO$_2$ diffusion within the leaf. Specific anatomical features, such as the arrangement and density of veins, mesophyll structure, and chloroplast positioning, directly affect the leaf's capacity to conduct water and facilitate gas exchange, thereby impacting the overall photosynthetic rate.
\end{itemize}

\textbf{Evolutionary Biology} \cite{ref37}
\begin{itemize}
    \item Example question: What is the molecular evidence for the early evolution of photosynthesis?
    \item Answer: Molecular evidence for the early evolution of photosynthesis includes the identification of photosynthesis genes from green sulfur bacteria (Chlorobium tepidum) and green nonsulfur bacteria (Chloroflexus aurantiacus). Phylogenetic analyses of these genes indicate that heliobacteria are closest to the last common ancestor of all oxygenic photosynthetic lineages and that purple bacteria are the earliest emerging photosynthetic lineage, thereby challenging previous conclusions based on ribosomal RNA and heat shock proteins.
\end{itemize}

\textbf{Agronomy} \cite{ref38}
\begin{itemize}
    \item Example question: How can canopy architecture optimization improve wheat photosynthesis and yield?
    \item Answer: Optimizing canopy architecture, such as promoting erect leaf angles and ensuring even light distribution throughout the canopy, can improve light interception and photosynthetic efficiency. This leads to more effective use of light energy and can enhance biomass production and yield in wheat crops.
\end{itemize}

\textbf{Environmental Science} \cite{ref39}
\begin{itemize}
    \item Example question: What is the impact of climate change on plant photosynthesis?
    \item Answer: Climate change alters temperature, precipitation, and soil moisture, thereby affecting plant photosynthesis. Rising temperatures can ease enzymatic limits on photosynthesis, although soil moisture deficits can limit CO$_2$ absorption by closing stomata.
\end{itemize}

\section{Results}
\subsection{Model Performance Evaluation}
During the model evaluation process, despite the impressive performance of GPT-4o, a significant performance difference was observed between PRAG and the baseline model (GPT-4o) when answering photosynthesis-related academic questions (Fig. 2a). Compared with the baseline model combined only with RAG, the combination of RAG and prompt optimization further improved the response performance (Fig. 2b). This produced encouraging results within the PRAG model evaluation framework. For example, the performance improved across all five metrics required for scientific paper writing in both the training and test sets, achieving an average improvement of 8.7\% compared with the baseline model. Notably, PRAG showed a significant improvement in source transparency, which increased from 5.2 to 6.4 points, a 25.4\% enhancement, through precise citation references and text source documentation.

GPT-4o (evaluated in June 2024) has demonstrated state-of-the-art performance in reasoning and text generation \cite{ref19}. In this study, we employed GPT-4o by combining the RAG technique with system prompt strategies to develop a photosynthesis research assistant model. The RAG technique expanded the knowledge scope of the model by embedding photosynthesis research papers into a vector database (DB), thereby significantly improving information accuracy and response quality. Furthermore, the system prompts provide an academic tone, proper citation formatting, and specific guidelines, further improving the model's capabilities for scientific paper writing.

To continuously improve the quality of the model responses, we introduced iterative refinement using a self-feedback approach (Fig. 2c). This method allowed the model to receive its own feedback and use it for improvements, thereby ensuring consistent quality without direct human intervention. Through an automated feedback loop, the model continuously evaluates and enhances its performance by adapting to various questions and contexts to provide more relevant responses. Following 10 iterations, PRAG optimized the prompts through self-feedback, generating relevant responses to user questions without relying on human annotations or refined data. Ultimately, this approach enhanced user experience and increased the model's reliability.
\begin{figure}
    \centering
    \includegraphics[width=1\textwidth]{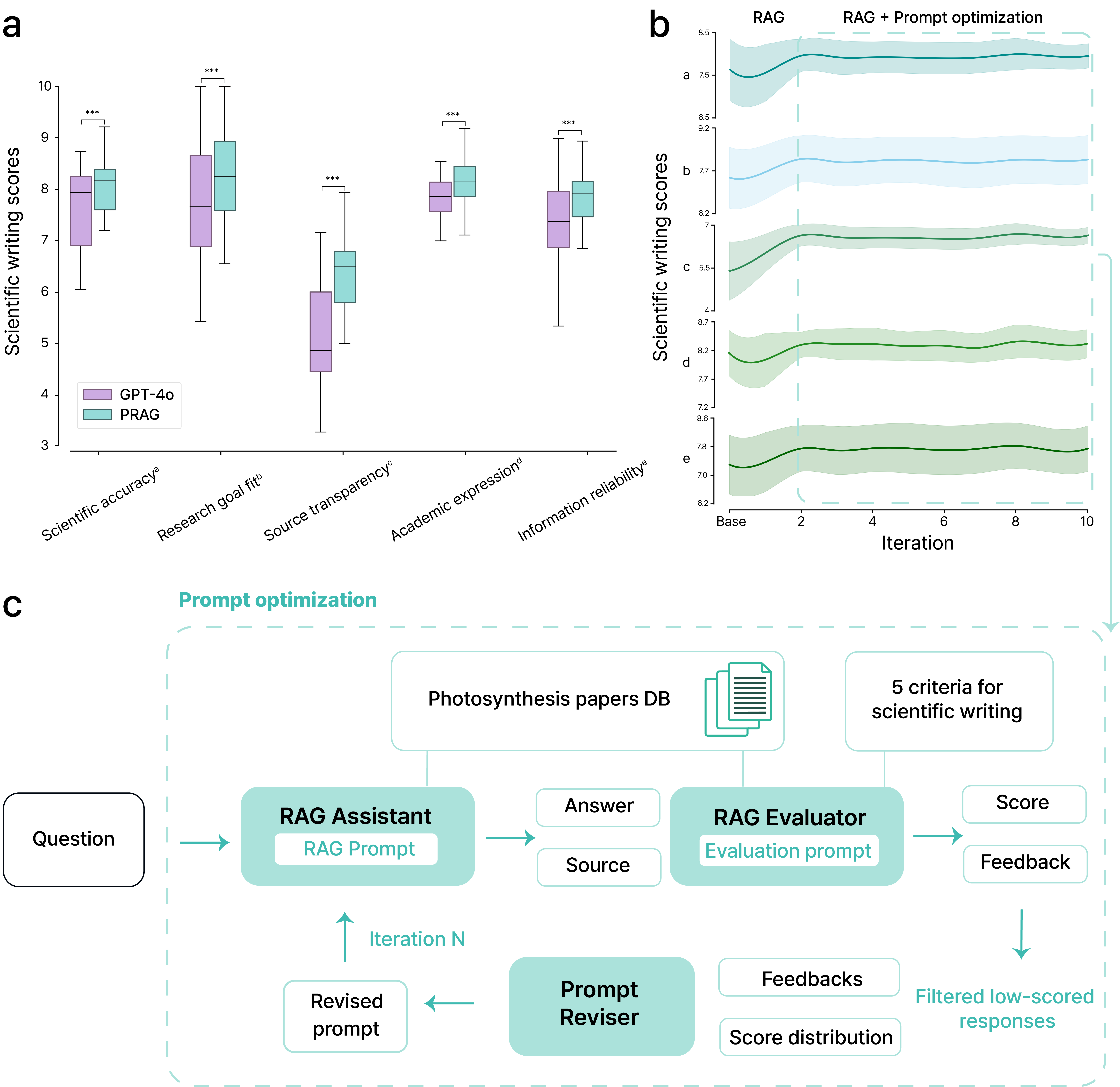}
    \captionsetup{
    labelfont=bf,
    labelsep=bar,
    font={small}
}
    \caption{Evaluation of the scientific writing performance of the photosynthesis research assistant PRAG and prompt optimization process. a, Responses of the improved model were evaluated based on five metrics required for scientific paper writing (scientific accuracy, alignment with research objectives, source transparency, scholarly tone, and information reliability). The evaluation was conducted using 150 training sets and 2,000 test sets out of 10,000 QA sets, thus reflecting the linguistic characteristics of QA sets extracted from photosynthesis research papers. The \textit{t}-test results showed significant differences between the two models across all metrics ($p < 0.001$). This indicates that the PRAG model performed significantly better overall compared to the GPT-4o model (*** indicates $p < 0.001$). b, Score change trends across the five evaluation metrics were analyzed as the RAG and prompt optimization processes were applied. The plot shows the score changes from the base model (Base) to the model with RAG only (1st iteration) and then with both RAG and prompt optimization (2nd iteration, Sky blue dashed line). The light background indicates the standard deviation (SD) at each iteration. c, Prompt optimization process: When a research question is input, the RAG Assistant generates an initial response based on the photosynthesis research database. The generated response and citation references are evaluated by the RAG Evaluator using five evaluation metrics. Feedback (scores and evaluations) is then delivered to the Prompt Refiner, where prompt adjustments are made to improve the response. Responses with low scores are filtered out, and a total of 10 iterations are performed to achieve optimization.}
    \label{fig:prag}
\end{figure}
\subsection{Research Text Visualization}
To demonstrate that PRAG can contribute to scientific insights and findings as a photosynthesis research assistant, we structured and visualized the response text of the model. First, 30 photosynthesis research papers were prepared for inclusion in the database, while 30 were excluded from the database and used as test papers. Next, we converted the core claims and hypotheses of these studies into questions. Subsequently, we saved PRAG's responses (discussions) to these hypothesis questions as 60 PDFs to create data that were comparative with the research papers.

\begin{figure}[htbp!]
    \centering
    \includegraphics[width=1\textwidth]{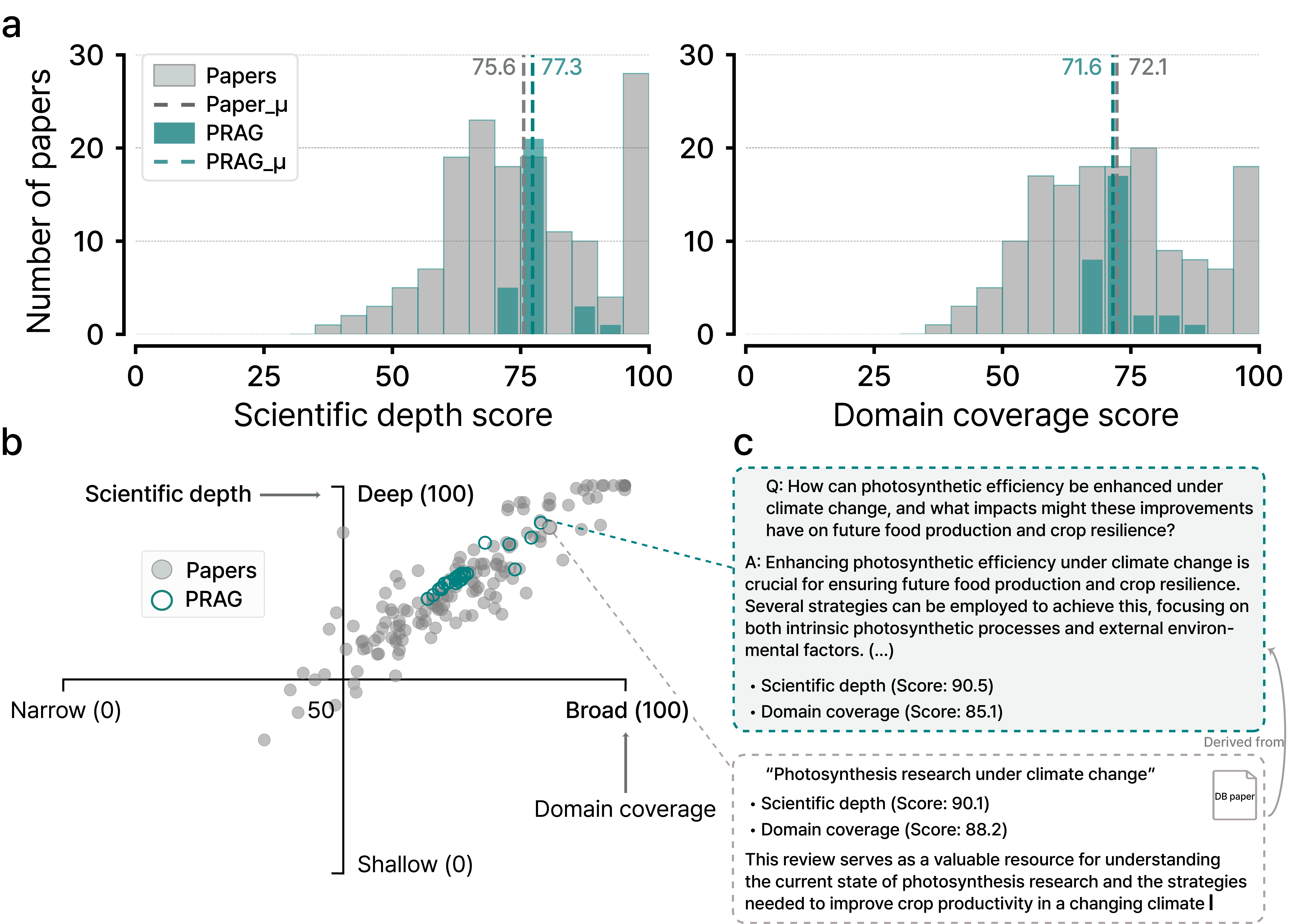}
    \captionsetup{
    labelfont=bf,
    labelsep=bar,
    font={small}
}
    \caption{Analysis of the scientific depth and domain coverage for photosynthesis research papers and PRAG discussion. a, Histogram showing the scientific depth (top-left) and domain coverage scores (top-right) for 150 photosynthesis research papers using the GPT-4o-mini model (gray color) and 30 PRAG discussions (teal color). b, Scatter plot depicting the relationship between scientific depth and domain coverage scores for photosynthesis-related texts. c. Highest-scoring PRAG discussion (teal dashed box), research paper text sample containing the corresponding hypothesis (gray dashed box), and evaluation details. The full set of PRAG discussions (PDFs) and code for the scientific text evaluation model are available at \url{https://github.com/PRAG-SNU}.
    }
    \label{fig:prag}
\end{figure}

To verify whether PRAG's discussions ensure scientific depth and domain coverage at the level of the cited literature, we evaluated the text of 150 photosynthesis research papers embedded in the vector DB based on the metrics of scientific depth and domain coverage (Fig.~3a). The x-axis represents the domain coverage score, which measures how well the research text integrates various disciplines related to photosynthesis, whereas the y-axis represents the scientific depth score, which indicates how deeply the research text addresses the scientific understanding of photosynthesis (Fig.~3b).

The evaluation results showed that the average scientific depth score for the 150 DB papers was 75.6, whereas that for PRAG was slightly higher at 77.3. In terms of domain coverage, DB papers scored 72.1, while PRAG scored slightly lower at 71.6. The highest-scoring discussion addressed photosynthetic efficiency under climate change by combining detailed biological mechanisms with environmental factors (Fig.~3c). Moreover, it explained how the overall system efficiency can be enhanced through gene-level manipulation and physiological adjustments based on interactions with external environments.

By effectively using network analysis techniques, the hidden patterns and knowledge in extensive text and data can be identified~\cite{ref20}. We utilized a flexible knowledge graph structure to organize the research paper text and PRAG discussions, and significant concepts and relationships were extracted to compare contextual relationships (Fig.~4). A Python-based PDF parser was combined with the NetworkX module to extract entities and relationships from the text and construct a knowledge graph.
\begin{figure}
    \centering
    \includegraphics[width=1\textwidth]{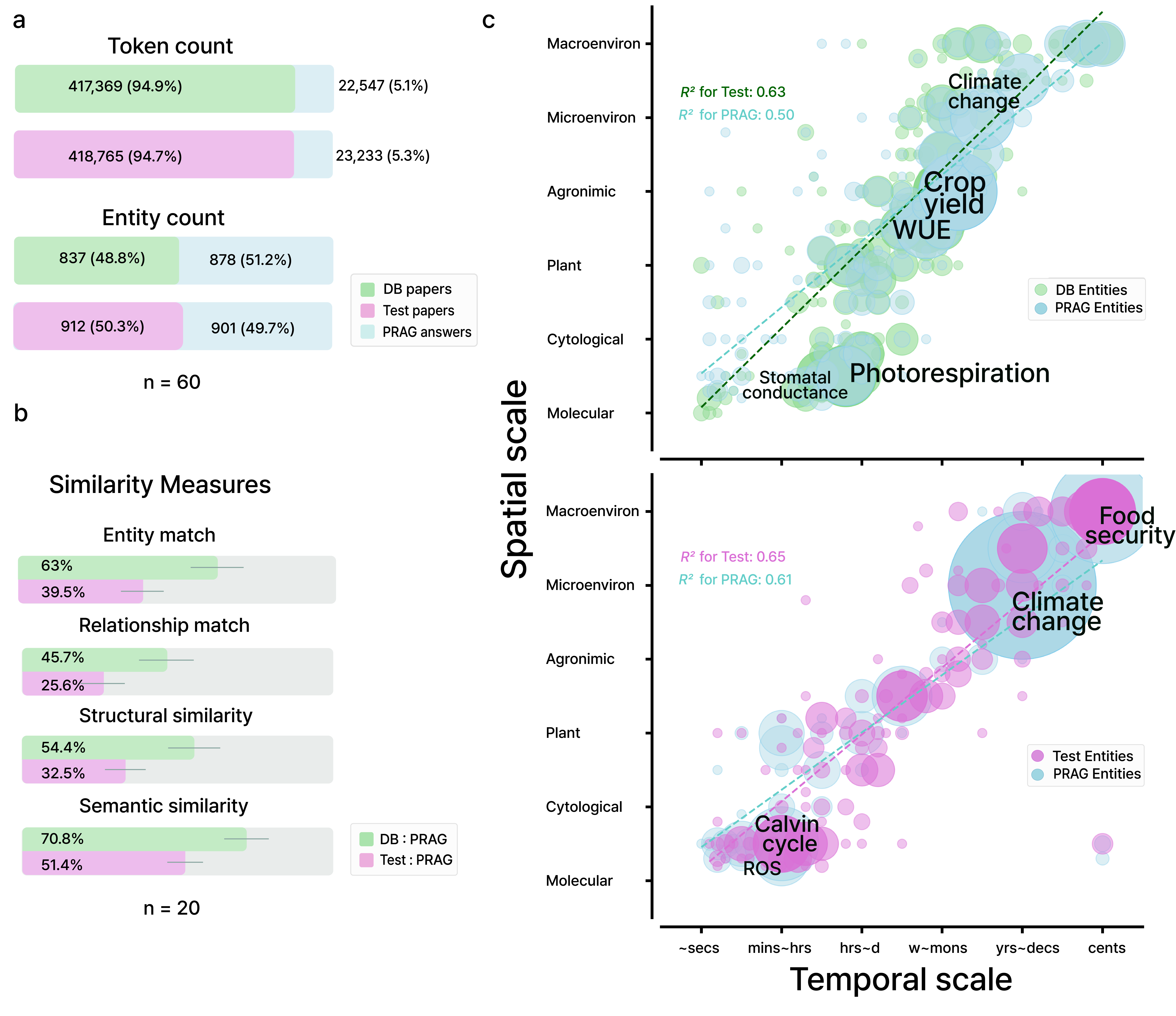}
    \captionsetup{
    labelfont=bf,
    labelsep=bar,
    font={small}
}
    \caption{Comparison of concepts and relationships in research papers and PRAG discussions using the knowledge graph. a, Comparison of the number of tokens and entities in database papers, test papers, and PRAG discussions. b, Similarity evaluation of the entity match rate, relationship match rate, structural similarity (average of entity and relationship matches), and semantic similarity (including similar entities and relationships, then averaged). c, Visualization of entity distribution across spatial and temporal scales, with entities extracted from scientific texts mapped across spatial scales (from the molecular to macro-environment level) and temporal scales (from immediate crop responses to long-term responses spanning centuries). The scientific text parser, knowledge graph construction code, and entity visualization code are available at \url{https://github.com/PRAG-SNU}.
    }
    \label{fig:prag}
\end{figure}
The number of tokens in PRAG's responses to hypotheses was 45,780, which represented approximately 5.4\% of the 839,134 tokens in the 60 DB and test papers. However, the extracted entities were similar, with 1,779 entities extracted for PRAG and 1,749 entities extracted for the papers (Fig.~4a). Compared to the DB papers, PRAG achieved an entity match rate of 63\%, relationship match rate of 45.7\%, structural similarity of 54.4\%, and semantic similarity of 70.8\%. Compared to the test papers, PRAG achieved an entity match rate of 39.5\%, relationship match rate of 25.6\%, structural similarity of 32.5\%, and semantic similarity of 51.4\%.

To visualize how the core concepts of the papers and PRAG were distributed across spatiotemporal dimensions, we mapped the spatial scale from the molecular level to the macroenvironmental level and the temporal scale from immediate plant responses to long-term responses spanning centuries. In the discussions of the DB papers and PRAG, entities such as 'Photorespiration,' 'Stomatal conductance,' 'Water use efficiency,' 'Crop yield,' and 'Climate change' frequently appeared, with correlations between data distributions of $R^2 = 0.63$ (DB papers) and $R^2 = 0.5$ (PRAG). In the discussions of the test papers and PRAG, entities such as 'ROS,' 'Calvin cycle,' 'Climate change,' and 'Food security' were common, with high correlations between data distributions of $R^2 = 0.65$ (test papers) and $R^2 = 0.61$ (PRAG).

\begin{figure}
    \centering
    \includegraphics[width=1\textwidth]{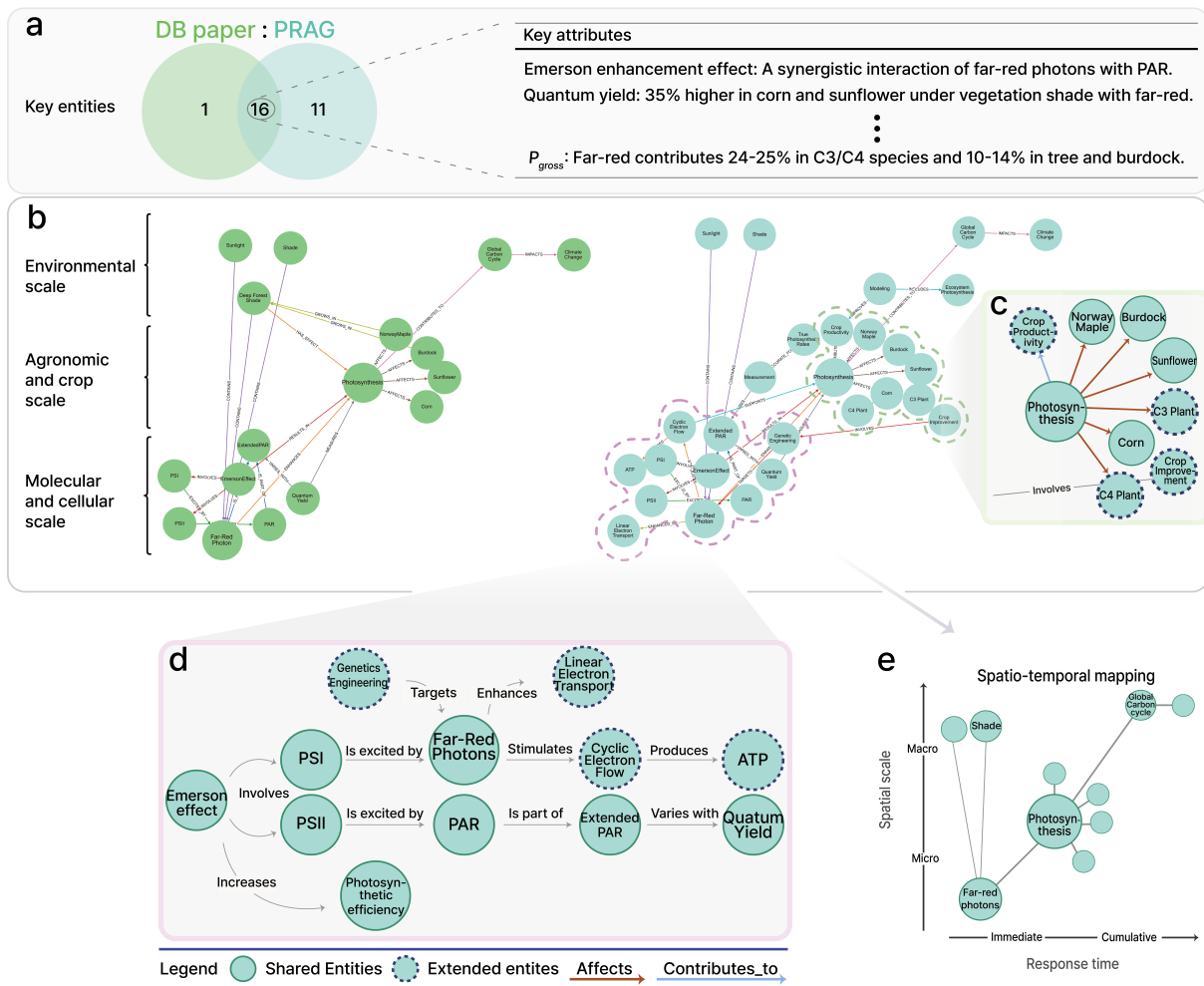}
    \captionsetup{
    labelfont=bf,
    labelsep=bar,
    font={small}
}
    \caption{Knowledge graph comparison between databased (DB) research papers and PRAG discussions. We used the knowledge graph to compare DB papers and PRAG discussions on the following questions: “How important are far-red photons in plant photosynthesis, and how do these photons affect crop photosynthetic efficiency and productivity under sunlight conditions?” a, Core entity diagram and key properties of shared entities in both DB papers and PRAG discussions. b, Knowledge graph comparison between DB papers and PRAG: Both graphs share key concepts related to the hypothesis, showing structural alignment at molecular and cellular levels to the macro-environmental scale. c, PRAG entities and relationships at the agricultural and crop scale. d, PRAG entities and relationships at the molecular and cellular scale: Dotted lines indicate concepts that were added by PRAG but omitted in the DB query. e, Mapping of knowledge graphs related to photosynthesis mechanisms across spatiotemporal scales. The knowledge graph illustrates the interactions related to photosynthesis at various spatiotemporal levels.}.
    \label{fig:prag}
\end{figure}

An example of the knowledge graph constructed from the discussions in the DB papers and PRAG is shown in Fig. 5. PRAG not only captured the core concepts of the DB paper regarding far-red photons but also elaborated on the key attributes and properties of each concept and their relationships, including specific numerical data. For instance, it explained how including the far-red photons in photosynthetically active radiation (PAR) influences quantum yield in maize and sunflower under vegetation shade conditions and described the contribution of far-red to leaf photosynthesis in specific plant species (corn, sunflower, Norway maple, and burdock) based on research data.

\begin{figure}
    \centering
    \includegraphics[width=1\textwidth]{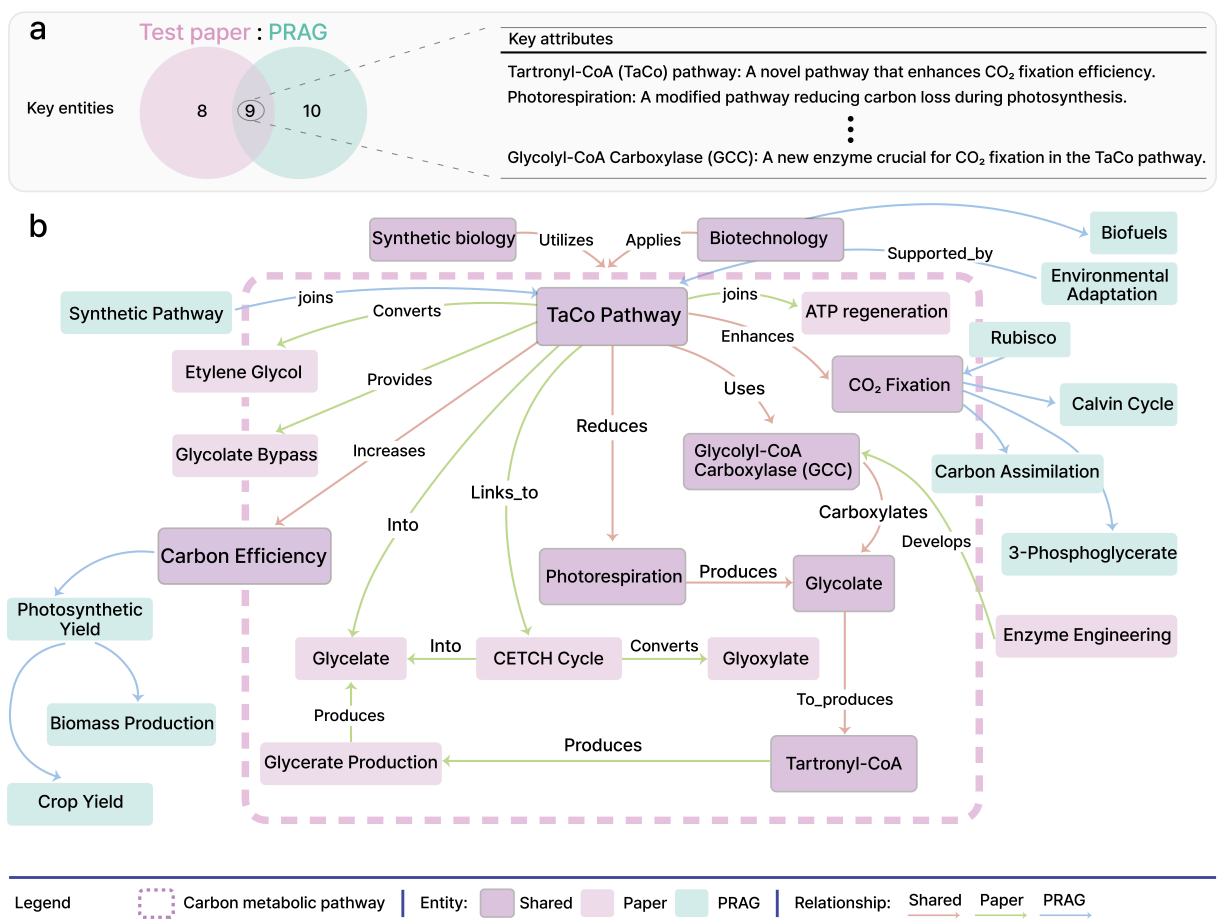}
    \captionsetup{
    labelfont=bf,
    labelsep=bar,
    font={small}
}
    \caption{Knowledge graph comparison between test papers and PRAG discussions. A knowledge graph was used to compare the test papers and PRAG discussions based on the following questions: “How does the development and implementation of the synthetic tartronyl-CoA (TaCo) pathway through the engineering of glycolyl-CoA carboxylase enhance the efficiency of carbon fixation in natural and synthetic systems, and what are the broader implications for improving photosynthetic yield and biotechnological applications under varying environmental conditions?” a, Core entity diagram and key properties of shared entities in both the test papers and PRAG discussions. b, Knowledge graph comparison between the test papers and PRAG. Both graphs share key concepts related to the hypothesis and depict the main interactions of the carbon metabolic pathway. The pink-colored dotted box represents the carbon metabolic pathway.}.
    \label{fig:prag}
\end{figure}

An example of the knowledge graph constructed from the discussions in the test papers and PRAG is shown in Fig. 6. PRAG addresses key concepts related to the synthetic TaCo pathway that are not included in the vector DB, such as glycolyl-CoA carboxylase, photorespiration, CO$_2$ fixation, biotechnology, synthetic biology, glycolate, and tartronyl-CoA, thereby expanding the discussion to include biotechnological mechanisms. The development of synthetic biological approaches to overcome limitations in carbon fixation pathways directly contributes to increased food production, environmental protection, and has significant impacts on the carbon-neutral bioeconomy. This test paper demonstrated that the TaCo pathway can be successfully reconstructed through enzyme engineering and applied for various biotechnological and agricultural uses. Similarly, PRAG inferred how the TaCo pathway connects to carbon fixation through glycolyl-CoA carboxylase (GCC), mitigates photorespiration, and enhances CO$_2$ fixation efficiency, thereby improving the overall photosynthetic rates. PRAG also mentions contributions to agricultural and biotechnological applications as well as biofuel production, thereby extending the discussion to a broader context.

\section{Discussion}
The research findings suggest that the performance improvement in response to photosynthesis research questions can be attributed to the capabilities of the LLM enhanced by the combination of RAG and system prompts. Although RAG alone has limitations in improving model performance, combining it with system prompts creates a synergistic effect that significantly enhances performance. This is because RAG effectively retrieves contextually relevant information from outside the domain, whereas the system prompts provide the model with a clear role and identity, thereby improving the quality of the output. We confirmed that the reasoning ability of the model related to photosynthesis research was enhanced through the prompt optimization process that combined a curated dataset with RAG and the clear instructions derived during that process. Various attempts have been made to utilize language models in the scientific field, including several inference-based LLMs \cite{ref21, ref22, ref23}. However, PRAG achieved comparable results without fine-tuning, thereby leveraging the excellent performance of the backbone model.

Scientific reasoning abilities include identifying research gaps, developing questions and hypotheses, identifying and classifying entities, and providing evidence through modeling, making claims, and performing evaluations \cite{ref24, ref25, ref26}. A well-structured scientific paper follows a C(Context)-C(Content)-C(Conclusion) structure, setting the context in the introduction, presenting content in the results, and drawing conclusions in the discussion \cite{ref27, ref28}. In this regard, PRAG effectively identified the achievements of prior research and remaining knowledge gaps in literature (Fig. SI.2) and reproduced the research logic from hypothesis (context) to conclusion (conclusion). Furthermore, PRAG discussions present sufficient scientific depth and are applicable across various academic fields covered by the academic DB papers (Fig. 3).

The field of plant science commonly uses bottom-up approaches that explain the effects of gene-level changes at the metabolomic, proteomic, and phenotypic levels and top-down approaches that explain specific phenotypes in the context of environmental factors \cite{ref26}. These processes enable the integration and understanding of complex biological phenomena related to photosynthesis. To understand how PRAG describes complex biological systems, we mapped it along with major concepts used in academic papers across various spatiotemporal scales. This demonstrated that the model does not remain at a localized scale but rather concretizes discussions from the macroenvironment to the molecular level (top-down approach) and extends time scales from immediate plant responses to long-term responses (bottom-up approach) (Fig. 4c; 5e).

Notably, through the knowledge graph, PRAG was able to secure key concepts related to the hypotheses using only approximately 5\% of the DB paper text, thus contributing to semantic expansion. This helps alleviate the cognitive burden of researchers and indicates that even very short texts can effectively summarize and integrate academic papers. By integrating the context of new citations, PRAG demonstrates its ability to provide an in-depth understanding of various topics, such as photosynthetic efficiency, agricultural productivity, crop resilience, and sustainable agriculture, and suggests future research directions. However, the expanded context may have led to differences in the correlations between data distributions by securing entities (concepts) across a wider range of spatiotemporal categories.

Moreover, improving photosynthetic efficiency is not limited to increasing plant productivity but rather represents a key research area linked to various industrial and environmental issues, such as clean energy production and climate crisis mitigation. In this context, enzyme engineering has the potential to improve carbon metabolic pathways, with PRAG offering practical insights from a comprehensive database of photosynthesis-related literature, thereby helping researchers formulate experimentally testable hypotheses. Researchers can draw inspiration from PRAG's predictions and designs to enhance their understanding, thereby securing a theoretical background to further promote research performance. This demonstrates how LLMs can support researchers in the theoretical and logical aspects of the scientific discovery cycle, particularly in the 'dry' phase where knowledge synthesis and hypothesis development occur. Just as systems biology advances scientific understanding through the complementary relationship between 'wet' (experiments) and 'dry' (computational analysis) approaches, our study shows that LLMs can effectively support researchers in the increasingly sophisticated knowledge synthesis and theoretical development process \cite{ref29}.

However, the database of the present study was limited to 150 papers due to resource and cost constraints and thus may not fully represent the diversity and depth of certain research areas. Additionally, a risk of bias toward specific researchers or groups was observed; therefore, future studies should include data from diverse sources to improve the model’s balance and accuracy. LLMs have limitations in interpreting numerical data. Therefore, incorporating multimodal features to analyze scientific charts and tables would be beneficial. Furthermore, ethical guidelines should be established to ensure fairness, accountability, and copyright protection to ensure the responsible use of LLMs.

The emergence of foundational models and LLMs presents a significant opportunity to advance AI development in plant science. Scientific breakthroughs in natural sciences demand clear, interpretable, and explainable domain knowledge \cite{ref30}. In particular, photosynthesis exhibits spatiotemporal complexity spanning from molecular to ecosystem levels, and this study demonstrated that LLMs can effectively support parts of the scientific discovery process in such complex photosynthesis research.

This study presents a novel approach to handling specialized photosynthesis knowledge by combining a knowledge database with LLMs. To address the challenge of evaluating semantic accuracy in text-based expert knowledge, we introduced a multifaceted validation methodology that incorporates entity matching, structural similarity, and spatiotemporal analysis through knowledge graphs. We demonstrated that PRAG extends beyond simple information retrieval and is able to comprehend the semantic relationships and contexts that exemplify expert knowledge. Moreover, it provides discussions at a level comparable to the original papers in terms of scientific depth and domain coverage using only a minimal portion of the source text. This suggests that LLMs can effectively support researchers by not only reducing their cognitive burden but also facilitating the complex process of knowledge integration and synthesis required in interdisciplinary research. Our multifaceted evaluation methodology serves as a framework for objectively validating LLM's contributions to the knowledge synthesis process in researchers' dry cycle and can be applied to future LLM research in other specialized fields.

\section{Code Availability}
The compiled code and files used in model development, including the API of the proprietary backbone model and the research data compiled into a database, are freely available for download at \url{https://github.com/PRAG-SNU}.

\section{References}
\vspace{-1cm}
\renewcommand{\refname}{}

\end{document}